\documentclass[a4paper]{article}

\usepackage{INTERSPEECH2022}
\usepackage{times}
\usepackage{amsmath}

\DeclareMathOperator*{\argmin}{arg\,min}
\usepackage{multirow}
\usepackage{subfigure}
\usepackage{subfig}
\usepackage{adjustbox}

\title{Why does Self-Supervised Learning for Speech Recognition Benefit Speaker Recognition?}
\name{Sanyuan Chen$^1$, Yu Wu$^2$, Chengyi Wang$^2$, Shujie Liu$^2$, Zhuo Chen$^2$, Peidong Wang$^2$, Gang Liu$^2$, Jinyu Li$^2$, Jian Wu$^2$, Xiangzhan Yu$^1$,  Furu Wei$^2$ }
\address{
  $^1$Harbin Institute of Technology, China \\
  $^2$Microsoft Corporation}
\email{
\begin{tabular}{c} 
\{yuwu1, t-chewang, shujliu, zhuc, peidongwang, liu.gang, jinyli, fuwei\}@microsoft.com \\
sychen@ir.hit.edu.cn,
jianwu@exchange.microsoft.com,
yxz@hit.edu.cn 
\end{tabular}
}

\begin{document}

\maketitle
\begin{abstract}
Recently, self-supervised learning (SSL) has demonstrated strong performance in speaker recognition, even if the pre-training objective is designed for speech recognition.  In this paper, we study which factor leads to the success of self-supervised learning on speaker-related tasks, e.g. speaker verification (SV), through a series of carefully designed experiments. Our empirical results on the Voxceleb-1 dataset suggest that the benefit of SSL to SV task is from a combination of mask speech prediction loss, data scale, and model size, while the SSL quantizer has a minor impact. We further employ the integrated gradients attribution method and loss landscape visualization to understand the effectiveness of self-supervised learning for speaker recognition performance.  

\end{abstract}

\section{Introduction}

Recently, self-supervised learning (SSL) has achieved the state-of-the-art results on a diverse array of downstream speech tasks \cite{wav2vec2, hsu2021hubert, zhang2021bigssl, superb, chen2021unispeech, chen2021wavlm, wang2021self}. Typical SSL methods either discriminate the correlated positive samples from the negative ones (e.g. wav2vec 2.0) \cite{wav2vec2}, or predict discrete pseudo-labels on the masked regions (e.g. HuBERT) \cite{hsu2021hubert}. Both methods try to implicitly learn short-time phonetic information from a huge amount of unlabeled speech, and mainly target at self-supervised learning for automatic speech recognition task (SSL4ASR). 

Due to the high correlation with phoneme units, it is straightforward to understand that SSL4ASR has the potential to drastically improve the speech recognition task.
Interestingly, SSL4ASR also achieves state-of-the-art performance on the speaker-related tasks, e.g. speaker verification (SV). For instance, WavLM \cite{chen2021wavlm} and BigSSL \cite{zhang2021bigssl} show the best performance on different partitions of VoxCeleb1 dataset \cite{nagrani2020voxceleb}, and the ensemble of WavLM model and Res2Net \cite{gao2019res2net,zhou2021resnext} ranks at the top position on VoxSRC 2021 speaker verification permanent leaderboard\footnote{https://competitions.codalab.org/competitions/34066\#results.} with the team name Strasbourg-Spk. 

In this work, our goal is to understand \textit{which factor leads to the success of SSL4ASR in speaker recognition}. 
Specifically, we try to answer the following questions:
\vspace{-1mm}
\begin{enumerate}
  \item \textit{Can supervised ASR model benefit the SV task?}
  \item \textit{How does SSL benefit the SV task?}
  \item \textit{What is the best SSL setup for the SV task?}
\end{enumerate}
\vspace{-1mm}
To this end, we carefully design and conduct a series of experiments to investigate what is the indispensable part of SSL. We also perform Integrated Gradients attribution analysis and loss landscape visualization to further understand the contribution of SSL to SV performance.

The main finding is three-fold as follows. 
 First, SSL4ASR models have significantly better transferability than supervised ASR models in an apple-to-apple comparison, indicating the SSL objective function is a key ingredient for achieving excellent transferability. 
 Second, the HuBERT style loss, mask speech prediction, is slightly better than other SSL losses, such as contrastive learning and Mean Squared Error (MSE) loss, while how to generate pseudo-labels has minor impacts on the performance of HuBERT style models. Even pre-training with simple clustering methods on raw inputs could provide good performance on the SV task. 
 Data augmentation proposed in WavLM \cite{chen2021wavlm} is very helpful, even if the pre-train data is scaled up to 94k hours. In addition, data scale and model scale have a strong correlation to  model transferability. 
 Third, our analysis shows that SSL models only learn speaker related knowledge with shallow layers in pre-training stage, while fine-tuning stage could unleash the full capability of the model. We observe that an SSL model could provide a wider optima in fine-tuning, which enables better resistance against small perturbation, stronger generalization capability, and easier SV model optimization.

\section{Background}

\begin{figure}[t]
	\centering
    \includegraphics[width=0.4\textwidth]{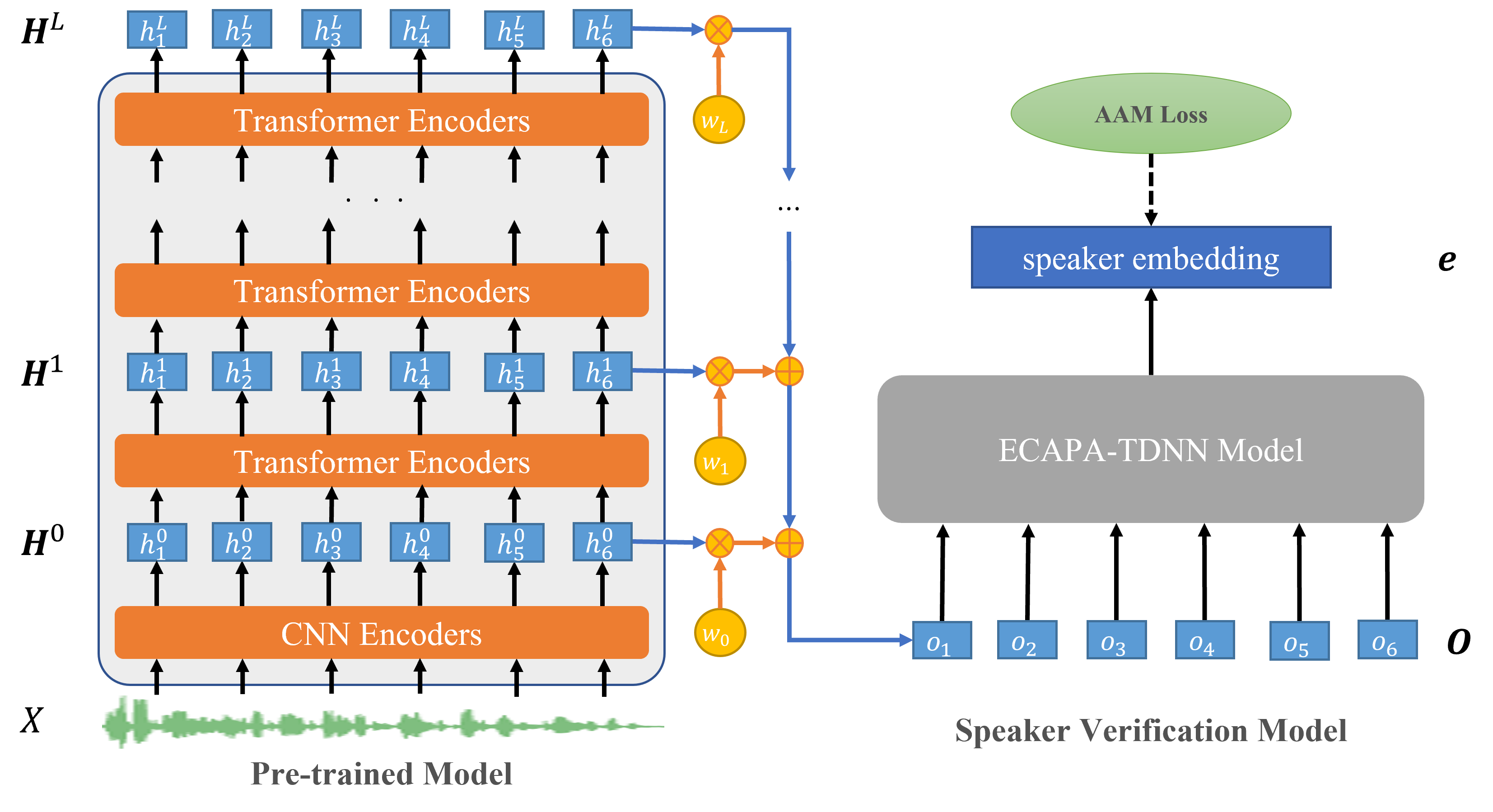}
    \label{fig:model}
	\vspace{-3mm}
    \caption{Self-supervised learning for speaker verification task.
    }
	\vspace{-3mm}
    \centering
\end{figure}

Self-supervised learning (SSL) has been shown to be an effective means of improving state-of-the-art results on SV task \cite{chen2021unispeech, chen2021large, chen2021wavlm}. 
A common practice is: 
We first optimize the pre-train model with SSL objective on the large-scale unsupervised data, then fine-tune the pre-trained model along with a downstream SV model on the annotated dataset. 

The typical SSL objectives are designed for automatic speech recognition task (SSL4ASR) by implicitly learning short-time phonetic information from unlabeled speech \cite{wav2vec2,hsu2021hubert}. 
Specifically, given a raw audio $X$, a latent representation $\mathbf{H}^0 = \{\mathbf{h}^0_t\}_{t=1}^T$ is obtained by a CNN feature extractor, where $T$ is the number of frames. Then the representation is fed to an $L$ layer Transformer model, yielding hidden states $\mathbf{H}^l = \{\mathbf{h}^l_t\}_{t=1}^T$, where $l$ denotes the $l$-th layer in the encoder. During pre-training, we employ the masked-based self-supervised learning methods to optimize the Transformer-based model. Before feeding the latent representation to the Transformer model, SSL methods first mask a proportion of them in some random frames, then minimize a variety of self-supervised objective functions based on the last layer hidden states output $\mathbf{H}^L$ in the masked regions.

During fine-tuning, we weighted average the hidden states of each layer to generate the output representation $\mathbf{O} = \{\mathbf{o}_t = \sum_{l=0}^L w_l \cdot \mathbf{h}_t^l\}_{t=1}^T$, where $w_l$ is a learnable weight for the hidden state of the $l$-th layer. 
Then we employ ECAPA-TDNN \cite{desplanques2020ecapa} as the downstream SV model following \cite{chen2021large}, and feed the output representation into the downstream model to generate the speaker embedding $\mathbf{e} = \text{ECAPA-TDNN}(\mathbf{o}_1, ..., \mathbf{o}_T)$. 
We use the additive angular margin (AAM) loss \cite{deng2019arcface} as the supervised objective function, and train the downstream SV model along with the pre-trained model for two stages. In the first stage, we optimize the parameters of the downstream model with the pretrained parameters fixed. In the second stage, we continue to optimize the parameters of the downstream model as well as the pre-trained model. In addition, we can also apply large-margin fine-tuning strategy and score calibration to further improve the speaker verification performance \cite{thienpondt2021idlab}.

\section{Why does SSL4ASR Benefit the SV task?}

\subsection{Can supervised ASR model benefit the SV task?}
\label{ssec:asr}

Given the similar modeling unit between SSL4ASR and supervised ASR models, it is a natural question whether the supervised ASR model can also benefit speaker verification task.
To verify this hypothesis, we compare the transferability of supervised ASR and SSL4ASR, both of which are trained on LibriSpeech 960h \cite{librispeech} and use the Transformer structure of HuBERT \cite{hsu2021hubert}.

The ASR model is trained with the Connectionist Temporal Classification (CTC) loss function \cite{ctc} in a supervised way. We use the character sequence as the target golden labels, and require the ASR model to predict the golden label $Y^{\text{golden}}$ given the hidden states of the last  encoder layer $\mathbf{H}^L$: $ \mathcal{L}_\text{CTC}  = - \log p_{\text{CTC}}(Y^{\text{golden}} | \mathbf{H}^L)$.
Spec-augmentation is also applied following \cite{specaug}.  

HuBERT, based on masked pseudo-label prediction loss, is selected as the SSL4ASR model for the comparison \cite{hsu2021hubert}. 
The pseudo labels are generated by iterative clustering.
At the first iteration, we conduct an offline clustering step on the MFCC feature of the input audio, where the clustering center of each frame is indexed as the pseudo label.
Then we use the hidden states $\mathbf{H}^L = \{\mathbf{h}^L_t\}_{t=1}^T$  to predict the embeddings $\mathbf{E} = \{\mathbf{e}_{y_t}\}_{t=1}^T$ corresponding to the pseudo labels $Y = \{y_t\}_{t=1}^T$ with the cross-entropy loss function in the masked regions $M$: 
\vspace{-2mm}
\begin{equation}
\vspace{-2mm}
\nonumber
\label{eq:hubert}
  \mathcal{L}_\text{HuBERT}  = \sum_{t \in M} - \log \frac{\exp (sim(\mathbf{h}^L_t \mathbf{W}, \mathbf{e}_{y_t})/\tau)}{\sum_{\hat{y}=1}^K \exp (sim(\mathbf{h}^L_t \mathbf{W}, \mathbf{e}_{\hat{y}})/\tau)},
\end{equation}
where $\mathbf{W}$ is the projection matrix, $sim(\cdot,\cdot)$ denote the cosine similarity function, $\tau$ is a pre-defined hyperparameter, and $K$ is the number of clusters.
Starting from the second iteration, we perform the offline clustering step with the hidden states extracted from the last iteration pretrained HuBERT model, and then train a new HuBERT model with the pseudo labels obtained by the new clustering centers.

We also use a random initialized Transformer model as a baseline to get rid of the effect of the additional parameters introduced by the pre-trained model and focus on the performance of different pre-training methods.

\begin{table}[t]
    \centering
     \small
    \caption{Transferability of supervised ASR and SSL4ASR
    }
    \vspace{-3mm}
    \label{tab:asr}
    \resizebox{0.7\columnwidth}{!}{
    \begin{tabular}{l|ccc} 
         \toprule 
         \multirow{2}{*}{Model} & \multicolumn{3}{c}{EER (\%)} \\
         & Vox1-O & Vox1-E & Vox1-H \\
         \hline
         FBank \cite{desplanques2020ecapa} & 1.01 & 1.24 & 2.32 \\
         \hline
         Random & 3.696 & 3.71 & 6.034 \\
         CTC & 1.159 & 1.256 & 2.434  \\
         HuBERT  & \textbf{0.84} & \textbf{0.879} & \textbf{1.726} \\
        
         \bottomrule
       
    \end{tabular}
    }
\end{table}

Table \ref{tab:asr} shows SSL4ASR model can provide a better representation than the handcrafted FBank feature, while the representations from the ASR model with CTC loss and the random initialized Transformer model are inferior to the FBank feature.
It indicates that the key to the success of SSL4ASR on SV task is neither the Transformer structure nor the  fine-tuning pipeline, but the self-supervised learning procedure.

\subsection{What is the best SSL objective for the SV task?}

Besides HuBERT, which is based on masked pseudo-label prediction loss, we also evaluate the transferability of wav2vec 2.0 \cite{wav2vec2} and Mean Squared Error (MSE) loss based pre-training method. It should be noted that all the three methods use the same mask setting proposed in HuBERT.  

MSE firstly calculates the FBank feature $\mathbf{F} = \{\mathbf{f}_t\}_{t=1}^T$ of the raw audio, then measures the mean square error between the FBank feature  and the linear projection of last layer hidden states output  $\mathbf{H}^L = \{\mathbf{h}^L_t\}_{t=1}^T$ in the masked regions $M$ as the objective function: $\mathcal{L}_\text{MSE}  = \sum_{t \in M} ||\mathbf{f}_t - \mathbf{h}_t \mathbf{W}||^2$.

Wav2vec 2.0 firstly discretizes the latent representation $\mathbf{h}^0_t$ of each masked timestep $t$ to the quantized latent representation $\mathbf{q}_t$, then uses the context representation $\mathbf{h}^L_t$  to identify the true quantized latent representation $\mathbf{q}_t$ out of a set of candidate representations $\hat{\mathbf{q}} \in \mathbf{Q}_t$ with contrastive loss function: 
\vspace{-2mm}
\begin{equation}
\vspace{-2mm}
\nonumber
\label{eq:wav2vec}
  \mathcal{L}_\text{wav2vec 2.0}  = \sum_{t \in M} - \log \frac{\exp (sim(\mathbf{h}^L_t \mathbf{W}, \mathbf{q}_t)/\tau)}{\sum_{\hat{\mathbf{q}} \in \mathbf{Q}_t} \exp (sim(\mathbf{h}^L_t \mathbf{W}, \hat{\mathbf{q}})/\tau)}.
\end{equation}

\begin{table}[t]
    \centering
    \small
    \caption{ SSL with different objective functions
    }
    \vspace{-3mm}
    \label{tab:loss}
    \resizebox{0.7\columnwidth}{!}{
    \begin{tabular}{l|ccc} 
         \toprule 
         \multirow{2}{*}{Model} & \multicolumn{3}{c}{EER (\%)} \\
         & Vox1-O & Vox1-E & Vox1-H \\
         \hline
        MSE  & 0.979 & 1.075 & 1.98 \\
         wav2vec 2.0  & 0.973 & 0.933 & 1.831 \\
         HuBERT & \textbf{0.84} & \textbf{0.879} & \textbf{1.726} \\
        
         \bottomrule
       
    \end{tabular}
    }
    \vspace{-3mm}
\end{table}

Table~\ref{tab:loss} demonstrates that all the three SSL methods can provide better representation than the FBank feature, which is attributed to the contextual speech representation learning from the masked speech. 
HuBERT achieves the best performance, indicating the better generalization and effectiveness of pseudo-label prediction loss than contrastive loss and MSE loss. 

\subsection{What is the best SSL quantizer for the SV task?}

Since HuBERT style loss is better than others, we explore the performance of different pseudo-label creation methods (quantizers) for HuBERT loss. 
Besides the MFCC Clustering and Hidden State Clustering introduced by HuBERT, we also experiment with the labels obtained by Random Projection \cite{chiu2022self}, VQ-VAE quantizers \cite{van2017neural}, and frame-phoneme alignment.

With random projection quantizer, we first extract the FBank features $\mathbf{F} = \{\mathbf{f}_t\}_{t=1}^T$ of the input audio, project $\mathbf{f}_t$ to the vector $\mathbf{A} \mathbf{f}_t$ with a random initialized matrix $\mathbf{A}$, and then find the closest vector from a set of random initialized vectors $\mathbf{V} = \{\mathbf{v}_i\}_{i=1}^K$, where $K$ is the vector (code) numbers. The pseudo label of $t$-th frame is defined as the index of the closest vector: $y_t = \argmin_i || \mathbf{v}_i - \mathbf{A} \mathbf{f}_t ||$.

With VQ-VAE quantizer, we first extract the FBank features $\mathbf{F} = \{\mathbf{f}_t\}_{t=1}^T$ of the input audio, and train a VQ-VAE model \cite{van2017neural} to reconstruct the FBank feature on LibriSpeech 960h \cite{librispeech}.
Given the latent variable $\mathbf{Z} = \{\mathbf{z}_t\}_{t=1}^T$ obtained by a $6$-layer Transformer-based encoder, we discretize it with the closest vector \{$\mathbf{\hat{z}}_t = \argmin_{\mathbf{v}_i} || \mathbf{v}_i - \mathbf{z}_t ||\}_{t=1}^T$ in a latent embedding space $\mathbf{V} = \{\mathbf{v}_i\}_{i=1}^K$, where $K$ is the embedding numbers,  and then reconstruct the features $z_q(\mathbf{F})$ with a $6$-layer Transformer-based decoder.
The training loss of VQ-VAE is to minimize the mean squared error between the reconstructed features and the input features, along with the difference between the encoded variable and the discrete variable: 
\vspace{-1mm}
\begin{equation}\small
\vspace{-1mm}
\nonumber
\label{eq:wav2vec}
  \mathcal{L}_\text{VQ-VAE}  = ||z_q(\mathbf{F}) - \mathbf{F}||^2 + ||\text{sg}[\mathbf{Z}] - \mathbf{\hat{Z}}||^2 + \beta ||\mathbf{Z} - \text{sg}[\mathbf{\hat{Z}}]||^2, 
\end{equation}
where $\text{sg}[\cdot]$ is the stopgradient operator and $\beta$ is a pre-defined hyperparameter. 
During inference,  the pseudo label of $t$-th frame is defined as the index of the discrete latent variables in the latent embedding space: $y_t = \argmin_i || \mathbf{v}_i - \mathbf{z}_t||$.

In addition, we also consider using the phoneme sequence of the input audio as the pseudo label to see if ASR-related pseudo label can benefit the SV performance. Here, we use force-alignment tool \cite{mcauliffe17_interspeech} to get the frame-phoneme pairs on LibriSpeech 960h data.

\begin{table}[t]
    \centering
    \caption{HuBERT style loss with different quantizers. 
    }
    \vspace{-3mm}
    \label{tab:quantizer}
    \resizebox{0.9\columnwidth}{!}{
    \begin{tabular}{l|ccc} 
         \toprule 
         \multirow{2}{*}{Model} & \multicolumn{3}{c}{EER (\%)} \\
         & Vox1-O & Vox1-E & Vox1-H \\
         \hline
        MFCC Clustering & 0.872 & 0.917 & 1.766 \\
        Hidden State Clustering  & 0.840 & \textbf{0.879} & 1.726 \\
        Random Projection (500 codes) & 0.899 & 0.95 & 1.775 \\
        Random Projection (8192 codes) & 0.883 & 0.903 & 1.675 \\
        VQ-VAE & \textbf{0.824} & 0.899 &	\textbf{1.655} \\
        Phoneme  & 0.867 & 0.918 & 1.776 \\
        
         \bottomrule
       
    \end{tabular}
    }
    \vspace{-3mm}
\end{table}

Table~\ref{tab:quantizer} shows that all the quantizers have similar performance on the speaker verification task. Even when we use the phone sequence as the pseudo label, which is irrelevant to the speaker information, we can still obtain a well-performed speaker verification model with the masked pseudo-label prediction SSL method.

\subsection{Large-Scale SSL on SV task}

Moreover, we also leverage the data augmentation and scale-up strategy to further strengthen the self-supervised learning for speaker verification task. 
Following WavLM \cite{chen2021wavlm}, we employ the masked speech denoising and prediction framework as the data augmented self-supervised learning method to improve pre-trained model robustness for complex acoustic environments and the preservation of speaker identity.
We also scale up unlabeled pre-training data to 94k hours of public audios \cite{chen2021wavlm}, including 60k hours of Libri-Light \cite{librilight}, 10k hours of GigaSpeech \cite{GigaSpeech2021}, and 24k hours of VoxPopuli \cite{wang2021voxpopuli}, and enlarge the model to 24 layer Transformers with 316M parameters.

\begin{table}[t]
    \centering
    \caption{Data and Model Scale Up. $^*$ means using large margin finetune and calibration 
    }
    \vspace{-3mm}
    \label{tab:scale}
    \resizebox{0.8\columnwidth}{!}{
    \begin{tabular}{l|ccc} 
         \toprule 
         \multirow{2}{*}{Model} & \multicolumn{3}{c}{EER (\%)} \\
         & Vox1-O & Vox1-E & Vox1-H \\
         \hline
        HuBERT 960h & 0.84 & 0.879 & 1.726 \\
        WavLM 960h & 0.777 & 0.829 & 1.629 \\
        HuBERT 94kh & 0.734 & 0.847 & 1.725 \\
        WavLM 94kh & 0.739 & 0.742 & 1.483 \\
        WavLM 94kh Large  & 0.505 & 0.579 & 1.176 \\
        WavLM 94kh Large$^*$  & \textbf{0.308} &  \textbf{0.462} & \textbf{0.906} \\

         \bottomrule
       
    \end{tabular}
    }
    \vspace{-3mm}
\end{table}

Table~\ref{tab:scale} shows that the data augmentation strategy used in WavLM can successfully benefit the self-supervised learning for SV task. The performance improvement would be more significant if we scale up the pre-training data to 94kh.
Thanks to the larger parameter capacity, the WavLM Large model can bring more than 20\% EER reduction compared to the WavLM Base model.
With the large-margin fine-tuning strategy and score calibration methods, the WavLM Large model can achieve  33.2\%, 27.1\%, and 8.8\% relatively EER reduction compared to the state-of-the-art supervised model  (Vox1-O: 0.461, Vox1-E: 0.634, Vox1-H: 0.993) \cite{zhao2021speakin} on all the three VoxCeleb1 trial lists.

\section{Discussion and Analysis}

\subsection{Contribution Attribution}

We employ the Integrated Gradients (IG) attribution method \cite{sundararajan2017axiomatic} to demonstrate how each layer of the pre-trained model contributes to the final SV performance.
Compared with method in \cite{chen2021unispeech, chen2021wavlm}, IG better models contribution estimation as it consider not only the layer weight, but also the magnitude of each layer's hidden states.
Specifically, given a well-trained downstream model $F(\cdot)$, the hidden states $\{\mathbf{H}^i\}_{i=1}^L$ extracted from all layers, and the corresponding learned weights $\{w_i\}_{i=1}^L$, the attribution score of $l$-th layer hidden states is assigned as: 
\vspace{-2mm}
\begin{equation} \tiny
\vspace{-2mm}
\nonumber
\label{eq:loss_ig}
  IG(\mathbf{H}^l) = \sum_{t,f} \left( \mathbf{H}^l \times \int_{\alpha=0}^1 \frac{\partial F ( \alpha (\sum_{i=0}^L w_i \cdot \mathbf{H}^i) )}{\partial \mathbf{H}^l} d\alpha \right), 
\end{equation}
where $\times$ denotes Hadamard product, 
and $\sum_{t,f} \left( \cdot \right)$ denotes the summation over the time and feature dimensions. The larger attribution score indicates the more importance of the corresponding hidden states. The summation of the attribution scores of all the hidden states indicates the final prediction of the SV model, i.e.,  $\sum_{l=0}^L IG(\mathbf{H}^l) = F(\sum_{l=0}^L w_l \cdot \mathbf{H}^l) - F(0)$.
Due to the intractability, we approximate $IG(\cdot)$ with the gradients summation as: 
\vspace{-2mm}
\begin{equation} 
\tiny
\vspace{-2mm}
\nonumber
\label{eq:loss_ig}
\displaystyle
  IG^{\text{approx}}(\mathbf{H}^l) = \sum_{t,f} \left( \mathbf{H}^l \times \frac{1}{K} \sum_{k=0}^K \frac{\partial F ( \frac{k}{K} (\sum_{i=0}^L w_i \cdot \mathbf{H}^i) )}{\partial \mathbf{H}^l} \right), 
\end{equation}
where $K$ is the number of approximation steps for computing integrated gradients. We set $K$ to 50 in our experiment.

\begin{figure}[tb]
	\centering
    \subfigure[Stage 1: we fix the pre-trained model and only train the downstream model. ]{\includegraphics[width=0.7\columnwidth, trim={1.00cm 0.68cm 8.0cm 2.8cm}, clip]{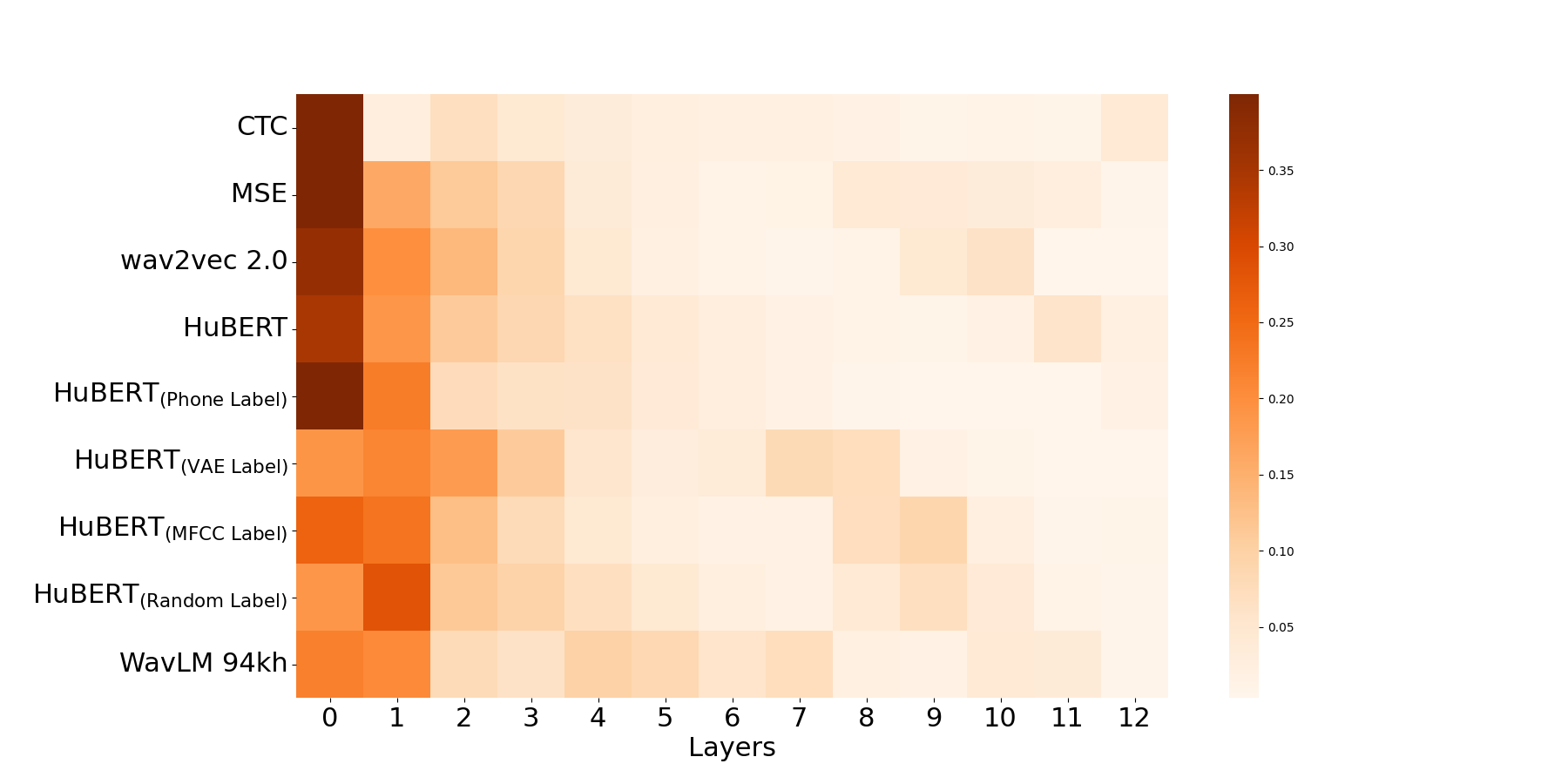}} \label{fig:ig1}
	\subfigure[Stage 2: we train both the pre-trained model and the downstream model. ]{\includegraphics[width=0.7\columnwidth, trim={1.00cm 0.68cm 8.0cm 2.8cm}, clip]{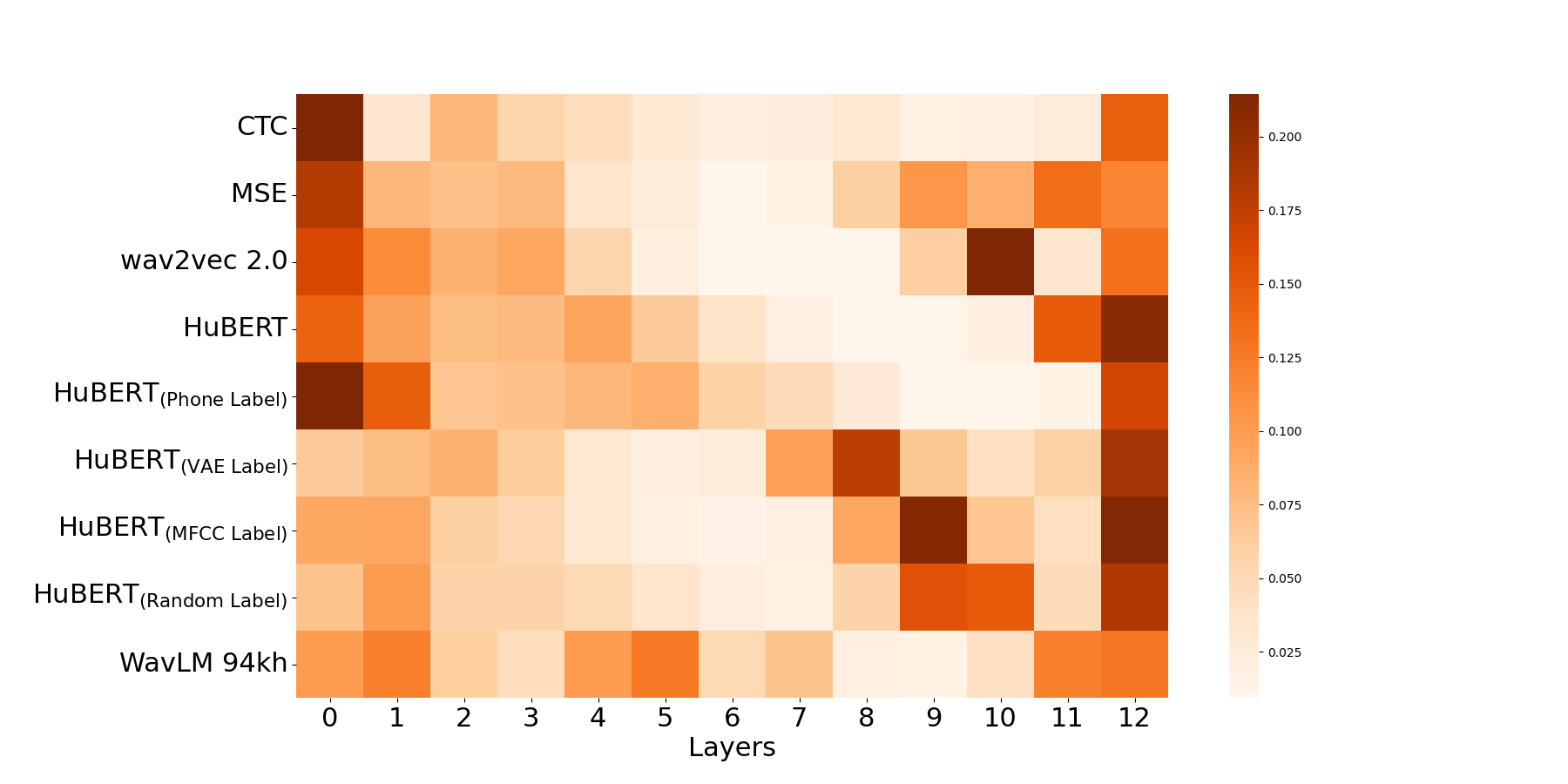}}  \\
	\vspace*{-3mm}
	\caption{Contribution attributed to each layer of each pre-trained model}
    \label{fig:ig}
	\vspace*{-5mm}
    \centering
\end{figure}

Figure~\ref{fig:ig} shows the contribution attribution from each layer of different pre-trained models. 
As for the first stage of fine-tuning, where we train the downstream model with the pre-trained parameters fixed, the contribution mostly comes from the output of the CNN feature extractor and the first encoder layer for all the pre-trained models. 
It indicates that only the shallow layers of pre-trained models learn the speaker-related information during the self-supervised learning procedure.
If the hidden states are extracted from the ASR model, which is supervised trained with CTC loss, only the latent feature extracted by CNN extractor contributes to the final prediction.
And if the hidden states are extracted from the SSL4ASR model, such as wav2vec 2.0 and HuBERT, the contribution is also dominated by the CNN extracted feature.
In contrast, if we pre-train HuBERT with data augmentation or the phoneme-independent quantizer, such as MFCC clustering or random projection, there are more contributions from the hidden states encoded by Transformer layers.  

As for the second stage of fine-tuning, we update the parameters of the downstream model as well as the pre-trained parameters. Since we unleash the full capability of the pre-trained model, the higher Transformer-based encoder layers can also learn to model the speaker information with the SV training objective, and make more contribution to the final prediction than in the first stage, leading to better speaker verification performance .

\subsection{Loss Landscape Visualization}

To better understand how self-supervised learning benefits the SV task, we visualize and compare the two-dimensional loss landscapes along with the optimization trajectories of different SV models.
For better comparison of different input features, we plot the parameters of the downstream models, and the optimization trajectories in the first fine-tuning stage where the pre-trained parameters are kept frozen.

Following \cite{li2018visualizing, hao2019visualizing}, we first define the origin and two axes of the loss surface as the random initialized downstream model's parameters and two directions in the parameter space, respectively.
Then, we uniformly sample multiple points around the initialized parameters, and plot the training loss of the downstream model with the parameters of each sampled point and the input feature from the pre-trained model.

Let $\mathbf{\theta}_0$, $\mathbf{\theta}_1$  denote the random initialized parameters and well-trained parameters of the SV downstream model respectively, we can define one of the axes as the optimization direction $\mathbf{\delta}_1 = \mathbf{\theta}_1 - \mathbf{\theta}_0$.
The other axis is set as a random direction $\mathbf{\delta}_2 = \mathbf{\theta}_2 - \mathbf{\theta}_0$, where $\mathbf{\theta}_2$ is the randomly generated parameters. Due to the high-dimensional parameter space, experimental results confirm that the two axes $\mathbf{\delta}_1$ and $\mathbf{\delta}_2$ are divergent and orthogonal to each other.
Then, the 2-D loss surface can be plotted with the function: $f(\alpha, \beta) = \mathcal{L}(\mathbf{\theta}_1 + \alpha \mathbf{\delta}_1 + \beta \mathbf{\delta}_2)$
, where $\alpha, \beta$ are scalar values and $\mathcal{L}$ is the loss function of the SV model training.
For better visualization, we scale the second direction vector $\mathbf{\delta}_2$ to the same norm as the first one $\mathbf{\delta}_1$ by $\mathbf{\delta}_2 = \frac{||\mathbf{\delta}_1||}{||\mathbf{\delta}_2||} \mathbf{\delta}_2$, where $||\cdot||$ is the Euclidean norm. We set the range of $\alpha$ and $\beta$  to $[-2, 2]$, and uniformly sample 29 points for each axis. In addition, we also project the optimization trajectory of the SV downstream model onto the two-dimensional loss surface.
Specifically, for the parameters of the downstream model $\mathbf{\theta}^i$ at $i$-th training epoch, $\mathbf{\delta}^i = \mathbf{\theta}^i - \mathbf{\theta}_0$ denotes the optimization direction at the $i$-th epoch, we can calculate the cosine similarity between the optimization direction $\mathbf{\delta}^i$ and each of the projected directions $\mathbf{\delta}$ as $\text{cos}(\mathbf{\delta}^i, \mathbf{\delta}) = \frac{\mathbf{\delta}^i \cdot \mathbf{\delta}}{||\mathbf{\delta}^i||  ||\mathbf{\delta}||}$. 
Then, the corresponding projected point $(\alpha^i, \beta^i)$ in the 2-D loss surface of $\mathbf{\theta}^i$ can be calculated as: $\alpha^i = \frac{||\mathbf{\delta}^i||}{||\mathbf{\delta}_1||} \text{cos}(\mathbf{\delta}^i, \mathbf{\delta}_1) = \frac{\mathbf{\delta}^i \cdot \mathbf{\delta}_1}{||\mathbf{\delta}_1||^2}, \beta^i = \frac{\mathbf{\delta}^i \cdot \mathbf{\delta}_2}{||\mathbf{\delta}_2||^2}$.

\begin{figure}[tb]
	\centering
    \includegraphics[width=0.14\textwidth]{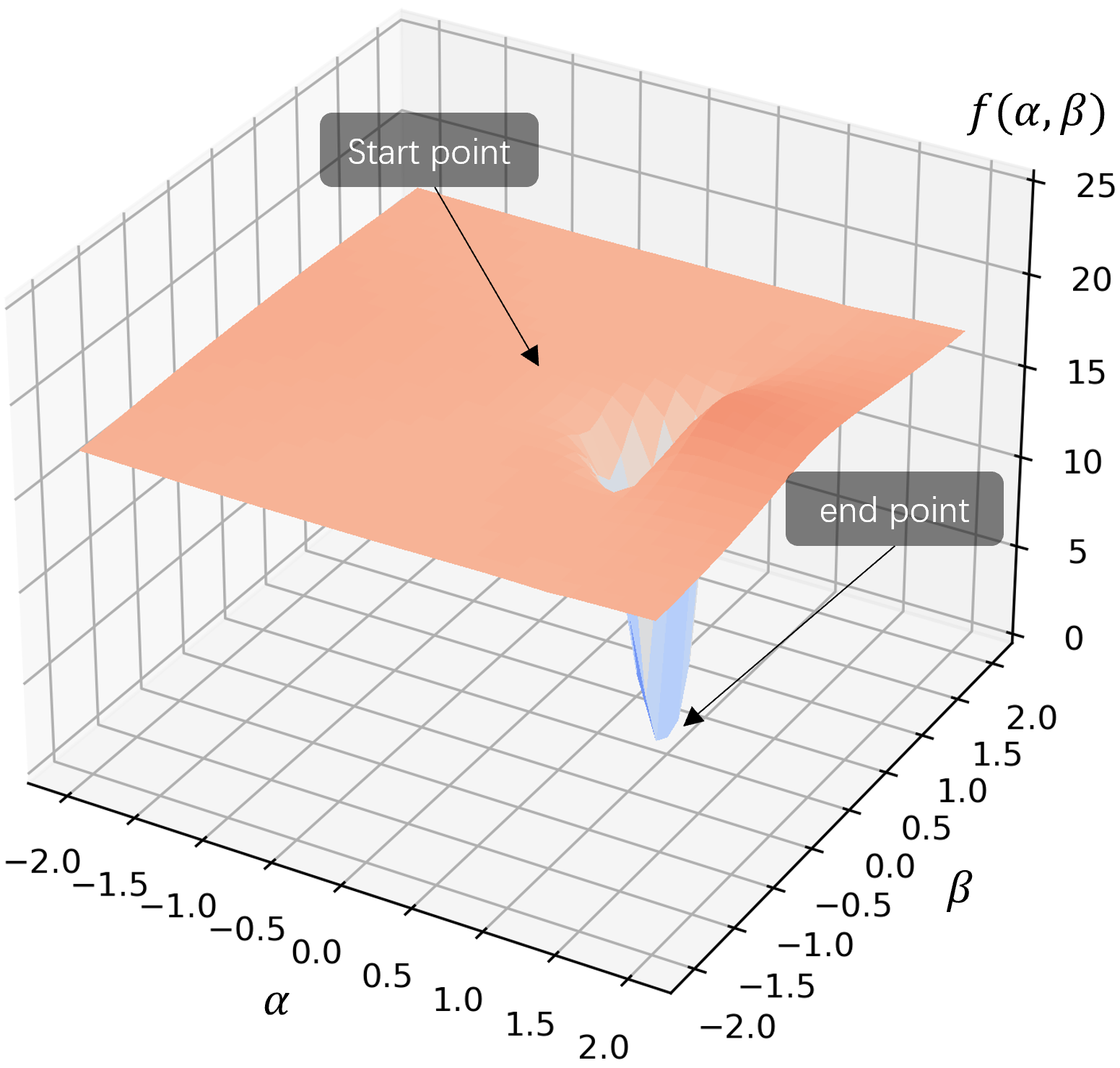}
	\includegraphics[width=0.14\textwidth]{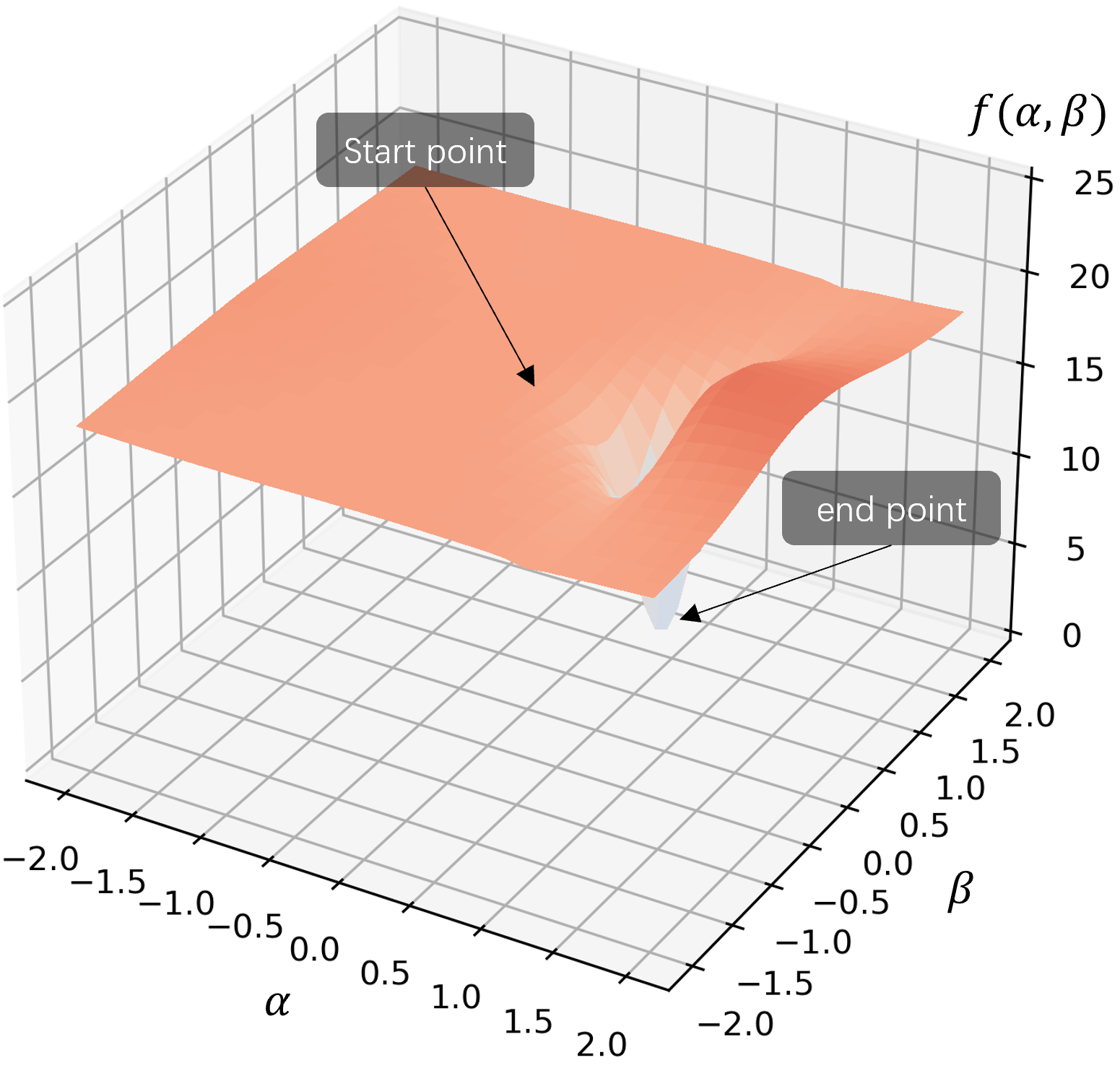}
	\includegraphics[width=0.14\textwidth]{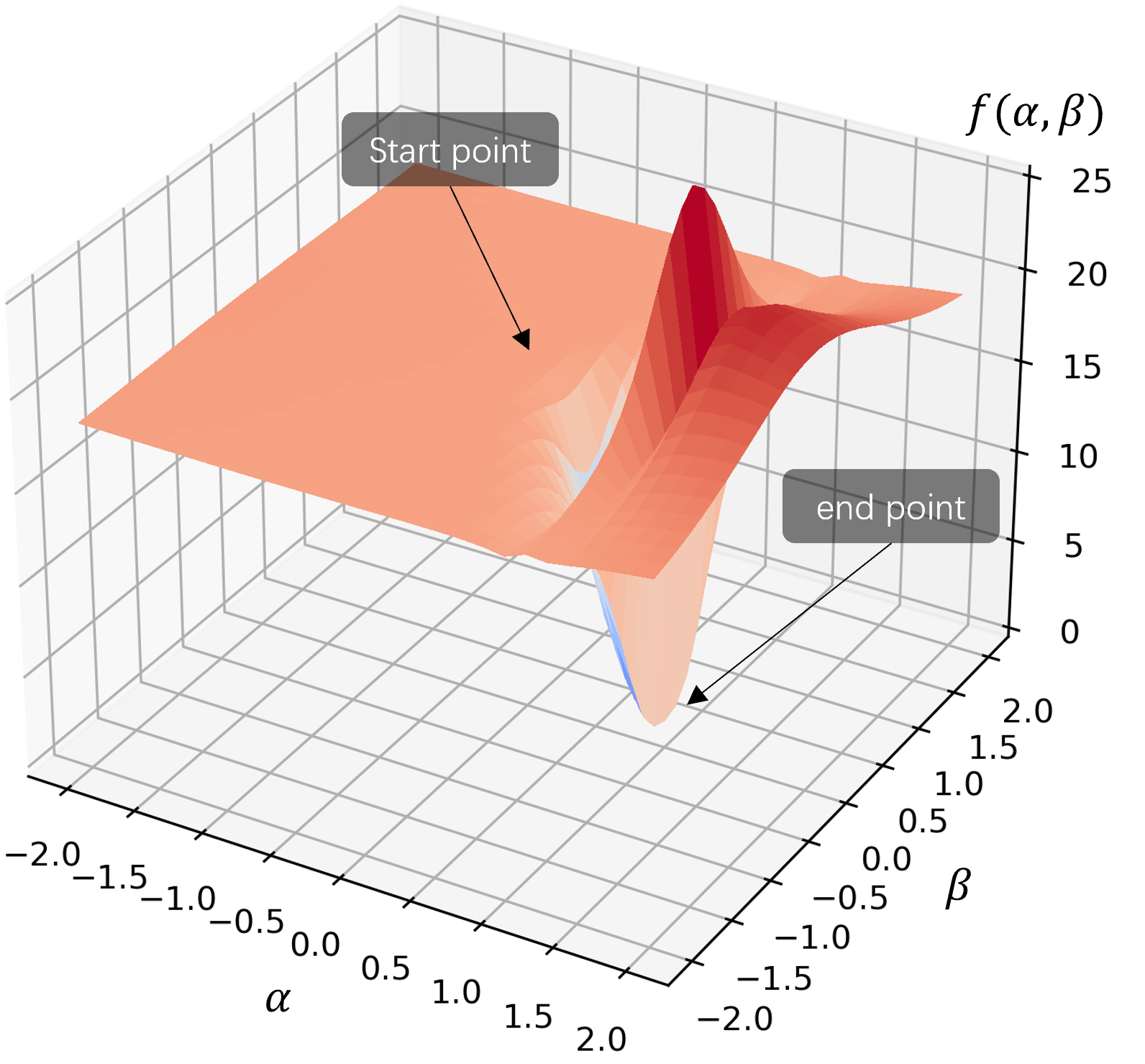}
	\includegraphics[width=0.02\textwidth]{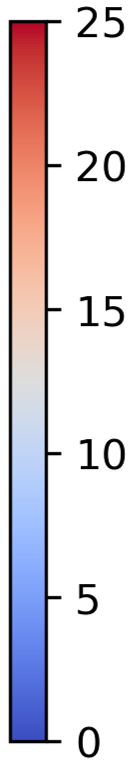} \\
	\subfigure[FBank feature]{\includegraphics[width=0.14\textwidth]{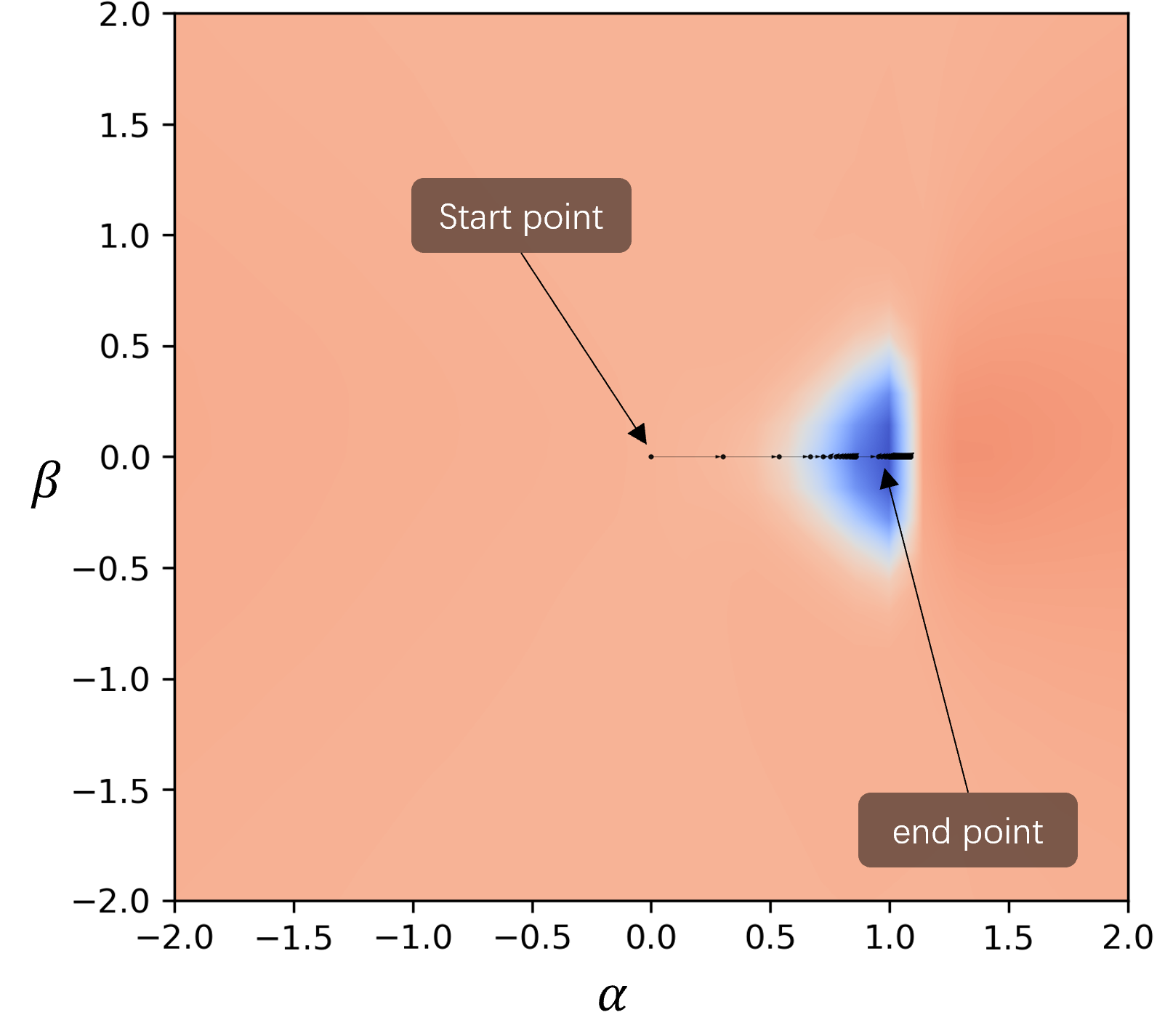}} 
	\subfigure[\footnotesize RI-WavLM]{\includegraphics[width=0.14\textwidth]{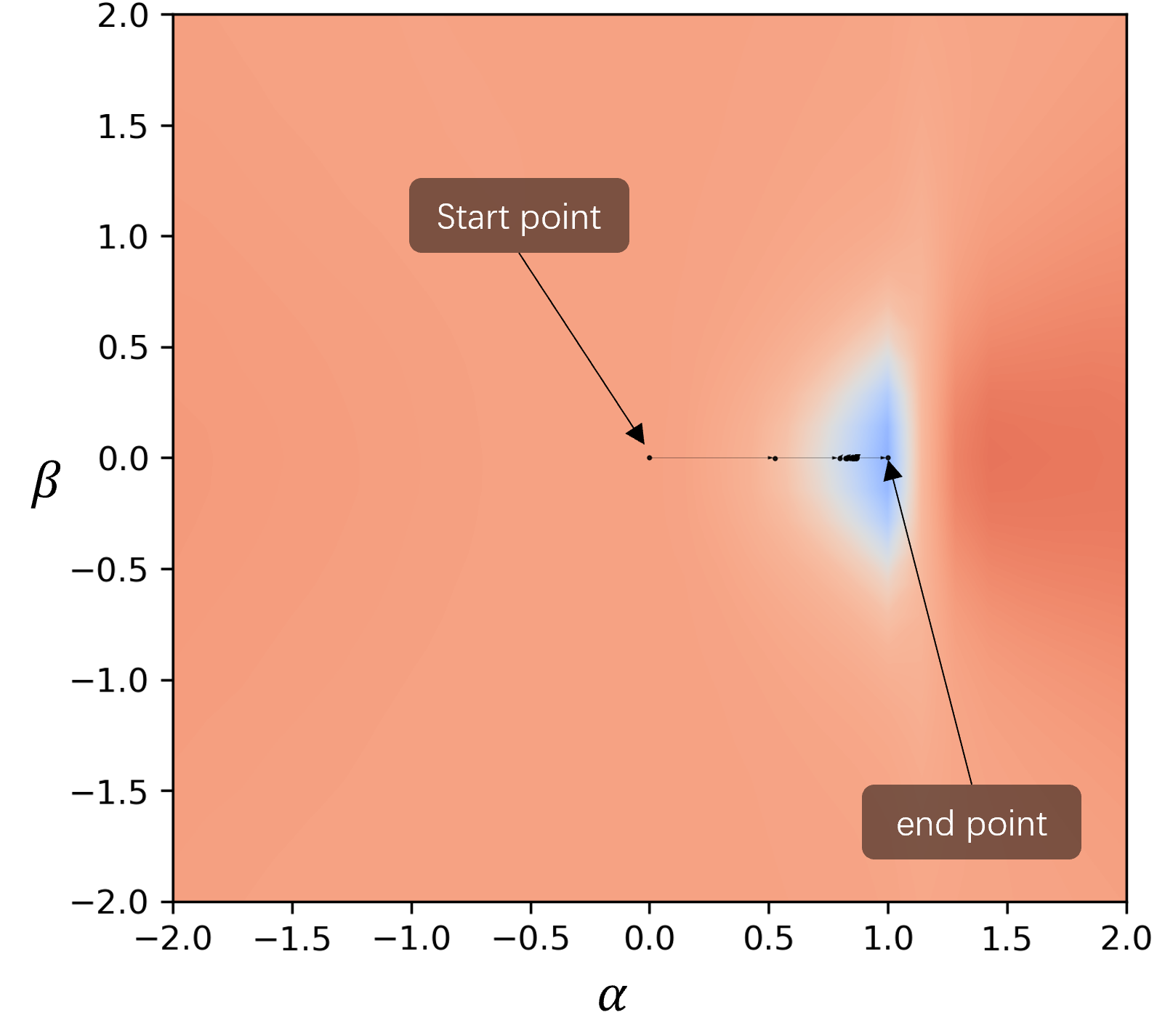}} 
	\subfigure[WavLM]{\includegraphics[width=0.14\textwidth]{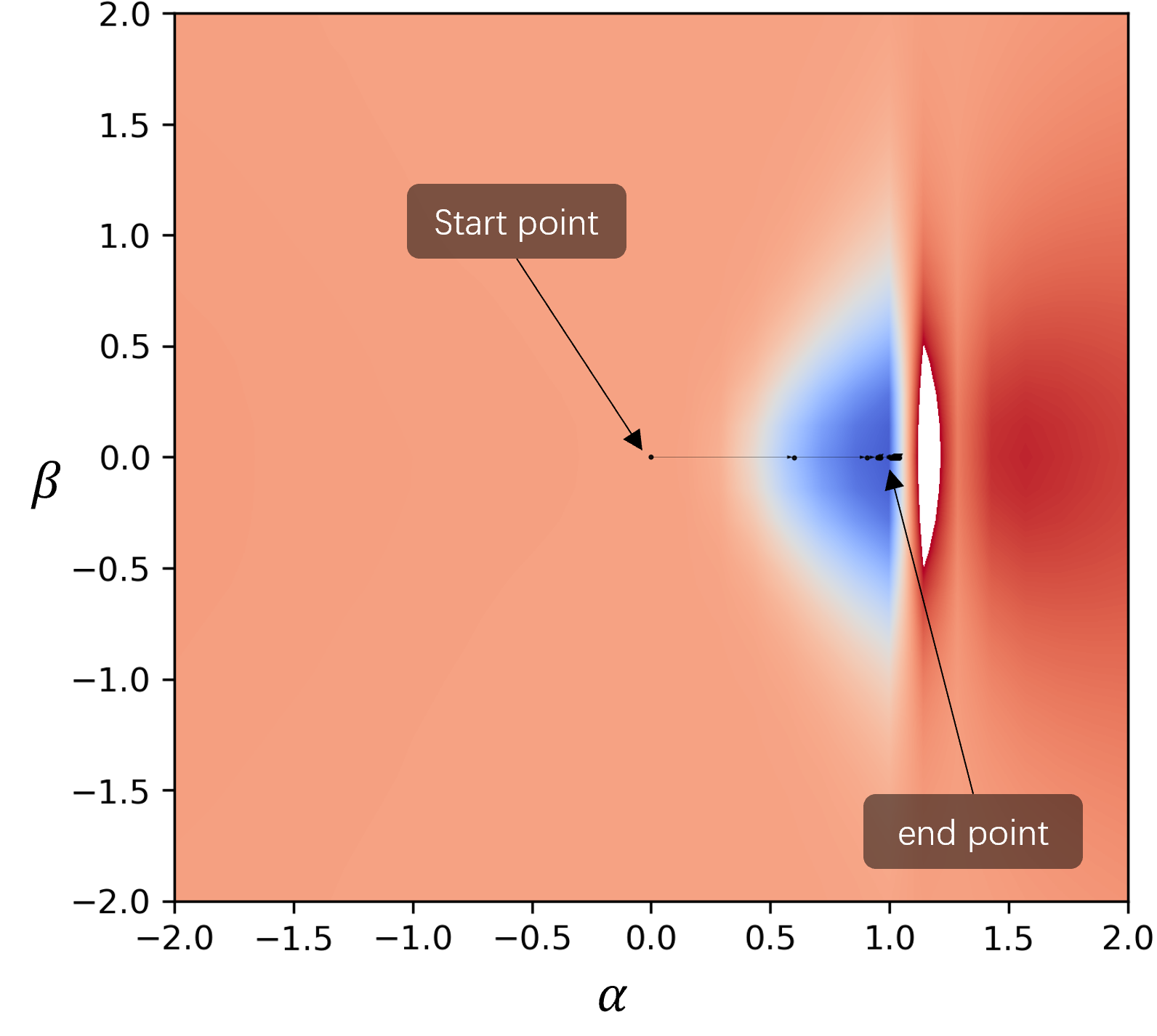}} 
	\includegraphics[width=0.02\textwidth]{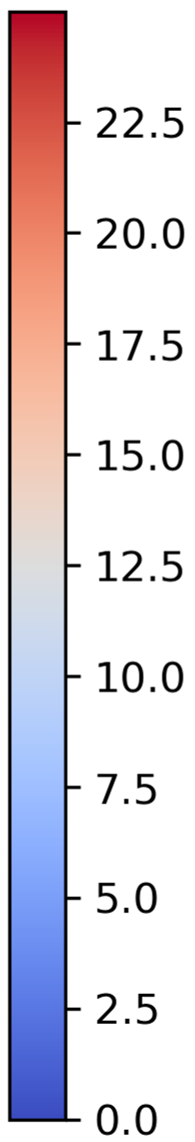} \\
	\vspace*{-3mm}
	\caption{Visualization of the loss landscape and optimization trajectories of SV model with different input features (FBank feature, Random-Initialized WavLM feature, and WavLM feature ). The figure below is a top view of the figure above.}
	\vspace*{-5mm}
    \label{fig:visual}
    \centering
\end{figure}

Figure~\ref{fig:visual} shows the visualization of speaker verification downstream model with different input features.
Compared with the FBank feature, we can find that the representation from the Random-Initialized WavLM model can provide a wider optima, which enables better resistance against some small perturbation, and leads to easier SV model optimization. 
However, without the self-supervised pretraining, the speaker verification model would stuck into a poor local minima with worse speaker verification performance.
With large-scale self-supervised learning, the pretrained WavLM representation can provide a  better initial point with a much broader and deeper optimum area. 
Even with some small disturbance, the WavLM input feature enables the downstream model to converge to the expected optimal region, and prevent it from skipping the optimal region with a steep loss hill.

\section{Conclusion} 
Our experimental results demonstrate that the self-supervised learning procedure is the key to the success on SV task. Among a variety of SSL methods, the masked pseudo-label prediction loss can provide the representation with best generalization capability on SV task, regardless of the pseudo-label creation methods.
We also show that data augmentation and model scale-up can further strengthen SSL for SV task.
Moreover, our  analyses show that two-stage fine-tuning can make use of the full capacity of SSL models, and that SSL models can facilitate the SV model optimization with a better initial point with a  broader and deeper optimum area.

\bibliographystyle{IEEEtran}

\bibliography{mybib}

\end{document}